\documentclass[letterpaper]{article} 
\usepackage{aaai2026}  
\usepackage{times}  
\usepackage{helvet}  
\usepackage{courier}  
\usepackage[hyphens]{url}  
\usepackage{graphicx} 
\usepackage{natbib}  
\usepackage{caption} 
\usepackage{algorithm}
\usepackage{algorithmic}
\usepackage{booktabs}
\usepackage{amssymb}
\usepackage{amsmath}

\usepackage{newfloat}
\usepackage{listings}
\DeclareCaptionStyle{ruled}{labelfont=normalfont,labelsep=colon,strut=off} 
\floatstyle{ruled}
\newfloat{listing}{tb}{lst}{}
\floatname{listing}{Listing}
\title{Human Motion Synthesis in 3D Scenes via Unified Scene Semantic Occupancy}
\author{
    Jingyu Gong\textsuperscript{\rm 1,\rm 2,\rm 3},
    Kunkun Tong\textsuperscript{\rm 1},
    Zhuoran Chen\textsuperscript{\rm 1},
    Chuanhan Yuan\textsuperscript{\rm 4},
    Mingang Chen\textsuperscript{\rm 3},\\
    Zhizhong Zhang\textsuperscript{\rm 1}, 
    Xin Tan\textsuperscript{\rm 1,\rm 2}\thanks{Corresponding Author.},
    Yuan Xie\textsuperscript{\rm 1,\rm 2}\\
}
\affiliations{
    \textsuperscript{\rm 1}School of Computer Science and Technology, East China Normal University, Shanghai, China\\
    \textsuperscript{\rm 2}Chongqing Key Laboratory of Precision Optics, Chongqing Institute of East China Normal University, Chongqing, China\\
    \textsuperscript{\rm 3}Shanghai Key Laboratory of Computer Software Evaluating and Testing, Shanghai, China\\
    \textsuperscript{\rm 4}College of Computer Science of Chongqing University, Chongqing, China\\

    \{jygong, zzzhang, xtan, yxie\}@cs.ecnu.edu.cn, \{51265901010, 10235102518\}@stu.ecnu.edu.cn,\\ 20241401019@stu.cqu.edu.cn, cmg@sscenter.sh.cn
}

\usepackage{bibentry}

\begin{document}

\maketitle

\begin{abstract}
Human motion synthesis in 3D scenes relies heavily on scene comprehension, while current methods focus mainly on scene structure but ignore the semantic understanding. In this paper, we propose a human motion synthesis framework that take an unified Scene Semantic Occupancy (SSO) for scene representation, termed SSOMotion. We design a bi-directional tri-plane decomposition to derive a compact version of the SSO, and scene semantics are mapped to an unified feature space via CLIP encoding and shared linear dimensionality reduction. Such strategy can derive the fine-grained scene semantic structures while significantly reduce redundant computations. We further take these scene hints and movement direction derived from instructions for motion control via frame-wise scene query. Extensive experiments and ablation studies conducted on cluttered scenes using ShapeNet furniture, as well as scanned scenes from PROX and Replica datasets, demonstrate its cutting-edge performance while validating its effectiveness and generalization ability. Code will be publicly available at https://github.com/jingyugong/SSOMotion.
\end{abstract}

\section{Introduction}

Understanding and simulating human behaviors in real 3D scenes has attracted lots of attention due to its wide application in robotics, games, and AR/VR~\cite{zhao2023synthesizing,huang2023diffusion,wang2024move,hwang2025scenemi}. 

Pioneers attempted to emulate human behavior and then synthesize human motion in given scenarios~\cite{starke2019nsm,cao2020long,wang2021synthesizing,wang2022towards}. 
Thanks to the advances in 3D scene understanding~\cite{qi2017pointnet,qi2017pointnet++,gong2021omni,gong2021boundary}, recent works can perceive the whole scene and provide guidance for scene-aware human motion synthesis~\cite{wang2021synthesizing,wang2022towards,tang2024unified}.
More recent works paid much attention on exploiting fine-grained scene structures, and grid sensors were widely utilized for its simplicity~\cite{lim2023mammos,lee2023locomotion,zhao2023synthesizing,jiang2024scaling,liu2024revisit,cen2024generating}. However, the scene semantic information, which is highly correlated with human behavior, cannot be effectively formulated and leveraged in motion synthesis.

This motivates us to explore a general representation comprising both scene semantics and structures, and finally we find that Scene Semantic Occupancy (SSO) can fulfill our requirements~\cite{cao2022monoscene,tong2023scene,ma2024cotr,chen2025youtube}. Nevertheless, directly performing feature extraction on the SSO is highly resource-consuming. In addition, semantic category definitions are dataset-dependent in current scene comprehension, making it challenging to achieve cross-dataset generalization.

\begin{figure}
    \centering
    \includegraphics[width=\linewidth]{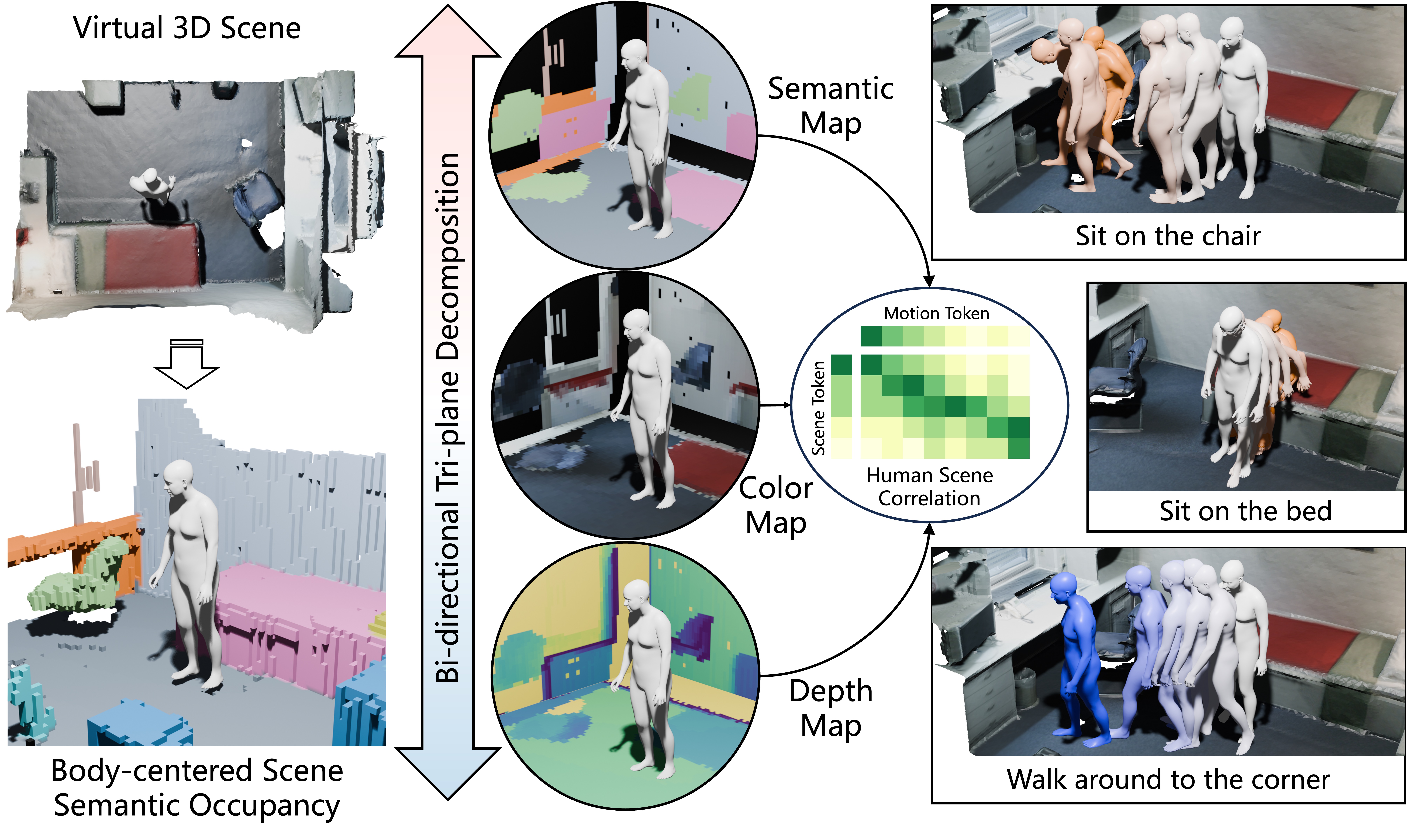}
    \caption{Illustration of human motion synthesis within 3D Scene Semantic Occupancy (SSO). We decompose the SSO into bi-directional tri-plane as a unified scene representation. Human scene correlation is modeled as control signal for instruction-aware motion synthesis in 3D scenes.}
    \label{fig:intro}
\end{figure}

Observing the inherent sparsity in scene occupancy representations, we propose to employ bi-directional tri-plane decomposition on it. We will first take grid sensors surrounding the human to obtain localized scene semantic occupancy, and then project the localized occupancy along the $\pm xyz$ axes to obtain semantic occupancy maps (Fig.~\ref{fig:intro} left). These semantic maps encapsulate body-centered semantic, depth, and color maps from diverse perspectives. Such low-dimension scene semantic occupancy can provide comprehensive semantic structure hints while substantially decreases computational demands. 

To take a unified semantic representation for the SSO, we take the CLIP text encoder~\cite{radford2021learning} for category feature embedding. Due to the repetitiveness of category information in semantic map and the high-dimension of CLIP feature, we decide to first take a shared mapping layer for semantic feature dimensionality reduction. Later, the compressed semantic features will be distributed into the semantic map for further scene comprehension. Thus far, we can derive the scene semantic structures for scene-aware motion synthesis with low computation overhead. 

These scene hints will further guide the human to achieve the goal following the instructions (Fig.~\ref{fig:intro} right). We choose to translate the instructions into target poses or positions via human population~\cite{hassan2021populating,zhao2022compositional}. Later, we model the correlation between 3D scene and goal-aware human motion via a frame-wise scene query, and then employ it for motion control in 3D scenes.

For performance evaluation, we have conducted experiments on cluttered scenes with ShapeNet~\cite{chang2015shapenet} furniture for primitive task evaluation. In addition, we follow the instructions to synthesize human motions in 3D scenes from PROX~\cite{hassan2019resolving} and Replica~\cite{straub2019replica} to show the cross-dataset performance. Extensive experiments and comprehensive ablation studies demonstrate the effectiveness and universality of the proposed method.

\section{Related Works}
\subsubsection{Human Motion Synthesis.} Motion synthesis has been researched for a long time~\cite{clavet2016motion,starke2019nsm}. As the technical development of motion capture and generative model, frontier researchers have continuously pursued generating more natural and realistic human motions~\cite{tevet2023human,amballa2025ls}. ACTOR~\cite{petrovich21actor} and TEMOS~\cite{petrovich2022temos} utilized a latent feature space for motion sequence, thus reducing accumulative error in motion synthesis. T2M~\cite{guo2022generating} estimated the motion length according to the text before final motion synthesis. MDM~\cite{tevet2023human} and MLD~\cite{chen2023executing} introduced diffusion model into motion synthesis and achieved performance boost. OmniControl~\cite{xie2024omnicontrol} utilized the joint hints to control the synthesized motion. MotionLCM~\cite{dai2024motionlcm} and MotionMamba~\cite{zhang2024motion} further improved the speed and efficiency for real-time application. Motion Anything~\cite{zhang2025motion} generated human motion according signals like text and music.

Thanks to the advancements in human motion synthesis, we can extend to scene-aware motion synthesis with easier control and higher motion naturalness.

\subsubsection{Scene-aware Motion Synthesis.} Scene-aware Human Motion Synthesis has attracted lots of attention recently due to its wide applications~\cite{cao2020long,hassan2021stochastic,hassan2023synthesizing,wang2024move,gong2026dip}. GAMMA~\cite{zhang2022wanderings} and DIMOS~\cite{zhao2023synthesizing} first learned a latent space for plausible motion and later trained policy models to control motion synthesis based on the latent space. SceneDiffuser~\cite{huang2023diffusion} designed a scene-based diffuser along with learning-based optimizer and planner to achieve the goal in 3D scenes. AMDM~\cite{wang2024move} introduced a two-stage generation framework and took scene affordance map as intermediate representation for human-scene interaction. SceneMI~\cite{hwang2025scenemi} utilized motion inbetweening with better key-frame control in scene-aware motion synthesis.

However, scene spatial semantics which is highly related to human motion is usually ignored by previous works. Thus, we decide to exploit the correlation between human behavior and scene semantics in this paper.

\subsubsection{Body-centered Scene Perception.} Scene perception has played a vital role in scene-aware motion synthesis. Pioneer~\cite{wang2021synthesizing} in scene-aware motion synthesis utilized PointNet~\cite{qi2017pointnet} to  extract scene feature. However, human motion is more related to body-centered local scene. PLACE~\cite{zhang2020place} calculated the distance between scene and human body to model the scene proximity. PSI~\cite{zhang2020psi} took scene semantics into consideration during static pose generation in 3D scenes. SAMP~\cite{hassan2021stochastic} and DIMOS~\cite{zhao2023synthesizing} designed virtual grid sensor to recognize the surrounded obstacles and avoid collision. Recent works~\cite{liu2024revisit,hwang2025scenemi} attempted to utilize space occupancy to represent the scene and provided scene perception for human motion synthesis.

Different from previous works, we introduce unified scene semantic occupany with bi-directional tri-plane decomposition as our scene representation. It can embed both scene structure and semantics in a lightweight way, and can generalize well to various scenes.

\section{Method}

\begin{figure*}[h!]
    \centering
    \includegraphics[width=\linewidth]{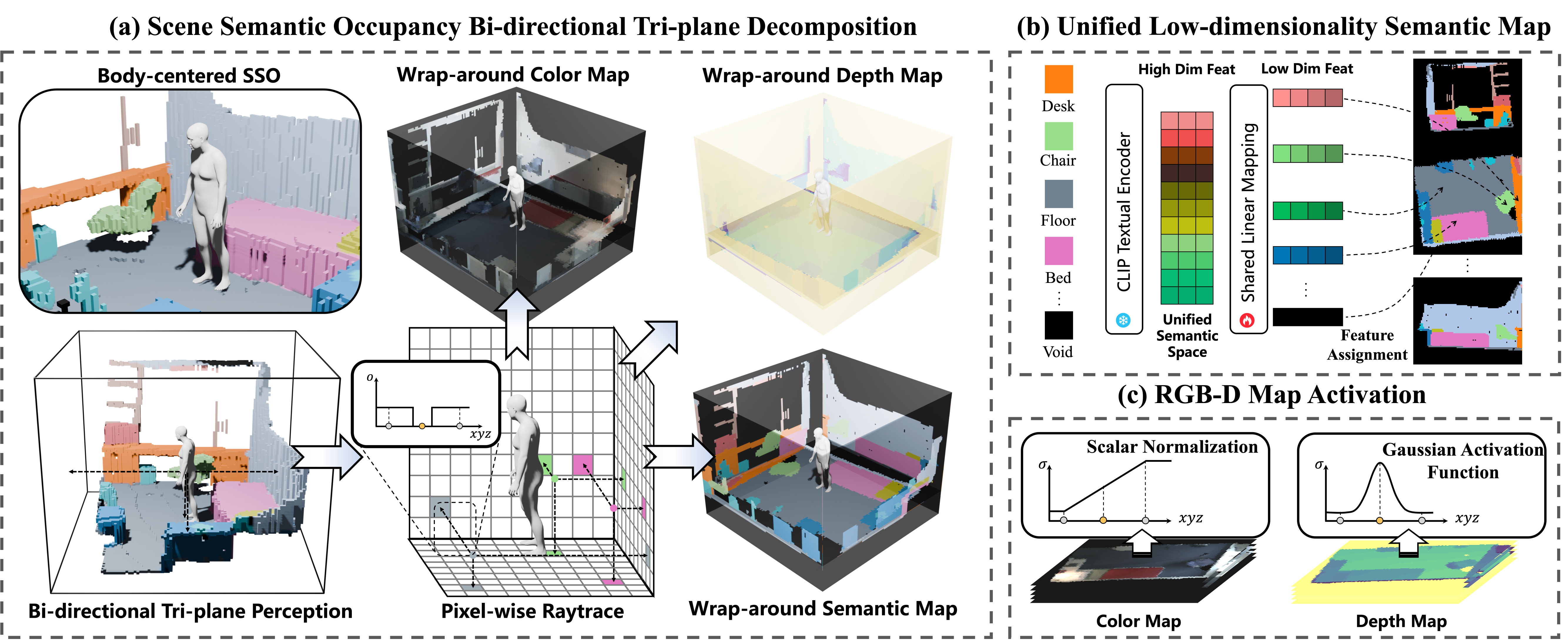}
    \caption{Pipeline of the Scene Semantic Occupancy perception (SSO). (a) presents the Bi-directional Tri-plane Decomposition of the SSO, where scene color, semantics and depth are perceived in body-centered coordinate. In (b), we map the semantic labels into a unified semantic space via the CLIP textual encoder and a shared linear layer. Then, the unified low-dimension semantic features will be scattered into the semantic map. (c) indicates the normalization functions for distance and color space.}
    \label{fig:pipeline}
\end{figure*}

\subsection{Preliminary}

\subsubsection{Human Motion.} Human motion can be treated as a sequence of body poses. Following previous works~\cite{wang2021synthesizing,zhao2023synthesizing}, we decide to take the parametric human model SMPL-X~\cite{pavlakos2019expressive} to represent the human pose. Within SMPL-X parameters, we mainly consider the global translation $\tau \in \mathbb{R}^3$, global orientation in axis-angle $\theta_{g}\in \mathbb{R}^3$, and body joint rotation in axis-angle $\theta_p=\theta_{j\in 1:21}\in\mathbb{R}^{63}$. Thus, the human pose can be represented by $P=\{\tau, \theta_{g}, \theta_{1:21}\}\in\mathbb{R}^{63}$. The human shape $\beta$ and hand pose $\theta_h$ remain invariant throughout the motion. The human mesh $\mathcal{M}$ and skeleton joints $\mathcal{J}$ can be directly derive from parameters aforementioned by SMPL-X human model $\mathcal{M},\mathcal{J}=\mathrm{SMPLX}(\tau,\theta_g ,\beta,\theta_{p},\theta_h)$.

\subsubsection{Scene Semantic Occupancy.} We decide to take scene semantic occupancy~\cite{cao2022monoscene} to represent a scene. As the occupied voxels in 3D scenes are sparse, we take a compact scene occupancy representation $\mathcal{S}\in{\mathbb{R}^{N\times 8}}$. Each element $\mathcal{S}_i$ in $\mathcal{S}$ indicate an occupied voxel, consisting of $xyz$ coordinate, $rgba$ color and semantic label $s$.

\subsubsection{Motion Diffusion Model.} For simplicity, we annotate any human motion as $x_0=\{P_s\}_{s\in1:S}$. The motion diffusion procedure will gradually add noise to the original motion
\begin{equation}
    q(x_t|x_{t-1})=\mathcal{N}(\sqrt{\alpha_t}x_{t-1},(1-\alpha_t)\mathbf{I}),
\end{equation}
where $\mathcal{N}$ indicates a normal distribution and $\alpha_{t\in{1:T}}$ are hyper-parameters. Finally, $x_T$ will approximate to $\mathcal{N}(\mathbf{0},\mathbf{I})$. In the reverse procedure, we will gradually denoise the human motion via $P(x_{t-1}|x_t)$. 

In our implementation, we supervise the network to directly predict the original motion given $x_t$ at any time step $t$ following MDM~\cite{tevet2023human}
\begin{equation}
    \hat{x}_0^{\phi}=\phi(x_t,t,c),
\end{equation}
where $c$ combines the action label $a$ and masked joints $\mathcal{J}_{1:S}$. During inference, we denoise the motion according to
\begin{equation}
P(x_{t-1}| x_t)=\mathcal{N}(\mu_t(\hat{x}_0^{\phi}(x_t),x_t),\tilde{\beta}_t\mathbf{I}).
\end{equation}
Here, $\tilde{\beta}_t = \frac{1-\bar{\alpha}_{t-1}}{1-\bar{\alpha}_t}\beta_t$, $\beta_t = 1-\alpha_t$, $\bar{\alpha}_t=\prod_{i=1}^t \alpha_i$, and $\mu_t(\hat{x}_0^{\phi}, x_t) = \frac{\sqrt{\bar{\alpha}_{t-1}}\beta_t}{1-\bar{\alpha}_t}\hat{x}_0^{\phi} + \frac{\sqrt{\bar{\alpha}_t}(1-\bar{\alpha}_{t-1})}{1-\bar{\alpha}_t}x_t$.

\subsection{Overview}

\subsubsection{Problem Definition.} This work attempts to synthesize human motion $\{P_s\}_{s=1:S}$ in 3D scenes $\mathcal{S}$ according to the textual instructions. The instruction contains a sequence of interaction sub-tasks (action and object pairs $\{(a_i,o_i)\}_{i=1:I}$). For each sub-task, the human may go somewhere or interact with the objects (\textit{e.g.} stools or tables) within scenes.

\subsubsection{Pipeline.} 
\begin{figure*}[h!]
    \centering
    \includegraphics[width=\linewidth]{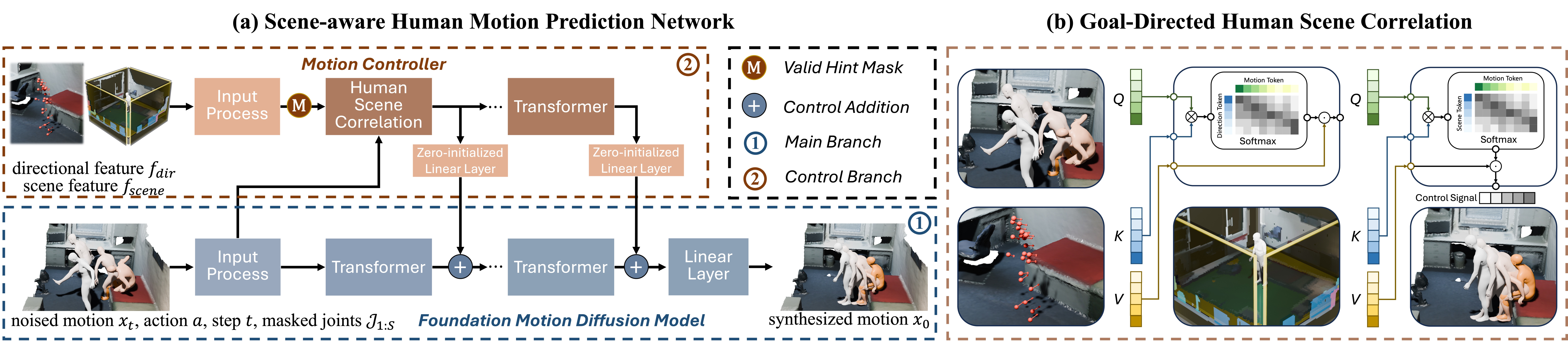}
    \caption{Overview of (a) the  network for instruction-aware human motion synthesis in 3D scenes and (b) the motion controller based on Goal-directed Human Scene Correlation.}
    \label{fig:network}
\end{figure*}

Given current task instruction $(a, o)$, we will synthesize future motion based on surrounding scene $\mathcal{S}_l$ and historical motion, as shown in Fig.~\ref{fig:pipeline} (a). Specifically, given any historical motion $\tilde{P}_{1:\tilde{S}}$, we decide to take the last $H=\min (\tilde{S}, H_{max})$ frames as our historical motion, where $\tilde{P}_{\tilde{S}-H+1}$ is treated as current human pose $P_1$, and $P_{1:H}=\tilde{P}_{\tilde{S}-H+1:\tilde{S}}$ will be utilize to control the motion synthesis. Specifically, we take the human joints $J_{1:H} = \tilde{J}_{\tilde{S}-H+1:\tilde{S}}$ as historical motion hint $f_h$. For body-centered scene perception, we take the localized scene semantic occupancy $\mathcal{S}_l$, and then decompose it into bi-directional tri-plane representation, as shown in Fig.~\ref{fig:pipeline} (b). Then, the bi-directional tri-plane semantic occupancy will be encoded into scene feature $f_{s}$ as shown in Fig.~\ref{fig:pipeline} (c). We further design a motion controller (Fig.~\ref{fig:network}) to synthesize future motion $\hat{P}_{1:S}$ based on historical motion $f_h$, current scene $f_s$ and future interaction goal $(a, o)$. Finally, the synthesized motion $\hat{P}_{1:S}$ will be updated to the historical motion $\tilde{S}_{1:\tilde{S}}$ with motion blending for overlapping $H$ frames.

\subsection{Semantic Occupancy Guidance}

\subsubsection{Bi-directional Tri-plane Decomposition.}
Given the global scene in a compact representation $\mathcal{S}$, we will first convert it into a sparse scene semantic occupancy $\mathcal{O}_g\in\mathbb{R}^{H_g\times W_g\times D_g\times 4}$, consisting of $rgb$ and semantic information. 

As shown in Fig.~\ref{fig:pipeline} (a), we will perceive the body-centered local scene based on current human status. Given any historical human motion $\tilde{P}_{1:\tilde{S}}$, we fetch the last $H$ frames $\tilde{P}_{\tilde{S}-H+1:\tilde{S}}$ as guidance for future scene-aware motion synthesis. We will take $\tilde{P}_{\tilde{S}-H+1}$ as the initial pose to perceive surrounding environments. Here, we can obtain human translation $\tau^{\mathcal{J}}$ and horizontal orientation $\theta_z^{\mathcal{J}}$ according to human skeleton joints $\tilde{\mathcal{J}}_{\tilde{S}-H+1}$. After that, we will arrange $H\times W\times D$ semantic occupancy sensors distributed evenly around human body $\mathcal{I}_l\in\mathbb{R}^{H\times W\times D\times 3}$. These sensors will be transformed into world coordinate through
\begin{equation}
    \mathcal{I}_g = R(\theta_z^{\mathcal{J}})\mathcal{I}_l + \tau^{\mathcal{J}},
\end{equation}
where $R(\cdot)$ indicates the rotation matrix. Later, these sensors will be taken to perceive the semantic occupancy. The local semantic occupancy can be represented by $\mathcal{O}_l\in \mathbb{R}^{H\times W\times D\times 4}$.

Last, the local semantic occupancy will be decomposed into bi-directional tri-plane. Specifically, $\mathcal{O}_l$ will be rendered along $\pm xyz$ axes, and obtain the semantic occupancy maps $\mathcal{O}_{yz},\mathcal{O}_{zy}\in \mathbb{R}^{H\times D \times 5}$, $\mathcal{O}_{zx}, \mathcal{O}_{xz}=\mathbb{R}^{W\times D\times 5}$, $\mathcal{O}_{xy},\mathcal{O}_{yx}\in \mathbb{R}^{H\times W\times 5}$. Each semantic occupancy map $\mathcal{O}_{ij}$ contains color $\mathcal{O}_{ij}^c$, depth $\mathcal{O}_{ij}^d$ and semantic label $\mathcal{O}_{ij}^{s}$  information. It's noteworthy, $\mathcal{O}_{yx}$ is not utilized in body-centered scene comprehension as ceiling won't influence human behavior in most cases.

\subsubsection{Unified Semantic Occupancy Representation.} As the semantic category sets are different for different 3D scene datasets, we need to take a unified semantic space for various category rather than utilizing one-hot embedding. It means we need to convert the semantic map $\mathcal{O}_{ij}^s$ into an unified semantic feature map $\mathcal{O}_{ij}^{sf}\in\mathbb{R}^{D_1 \times D_2 \times C}$, where $D_1,D_2$ indicates map shape, and $C$ indicates semantic feature dimension.

In this work, we choose to take the CLIP text encoder~\cite{radford2021learning} to embed the semantic category (Fig.~\ref{fig:pipeline} (b)). Due to the repetitive nature of semantics across most regions, performing feature extraction on high-dimensional CLIP semantic features $f_{clip}\in \mathbb{R}^{C_h}$ will result in significant computational redundancy. Thus, rather than directly transfer the semantic map into feature map, we choose to first employ a shared linear layer to reduce the dimensionality of all semantic category features. Then, the reduced-dimensional CLIP features $\tilde{f}_{clip}\in\mathbb{R}^{C_l}$ will be distributed into the semantic map, obtaining the unified semantic feature map $\mathcal{O}_{ij}^{sf}\in \mathbb{R}^{D_1\times D_2\times C_l}$. Based upon this, we can easily extract scene semantic feature $f_{sem}$ via a naive image encoder. 

For bi-directional tri-plane depth map $\mathcal{O}_{ij}^{d}$, as shown in Fig.~\ref{fig:pipeline} (c), we choose to focus more on nearby objects, as scene geometry in close proximity exhibits stronger correlation with human behaviors. Thus, we take a gaussian kernel as activation function to obtain the activated depth map
\begin{equation}
    \mathcal{O}_{ij}^{da} = \frac{1}{\sigma\sqrt{2\pi}}e^{-(\mathcal{O}_{ij}^d)^2/(2\sigma^2)},
\end{equation}
where nearby objects can be easily perceived. The scene geometry feature $f_{geo}$ can be directly extracted from it.

As for color map $\mathcal{O}_{ij}^c$, we simply normalize it to the range $[0,1]$, and derive the scene texture feature $f_{tex}$.

By now, the perceived scene feature $f_{scene}$ will combine the semantic, geometry, and texture feature
\begin{equation}
    f_{scene} = f_{sem}\oplus f_{geo}\oplus f_{tex},
\end{equation}
which will be utilized for motion control in 3D scenes.

\subsection{Motion Controller}

\subsubsection{Action Intention Cues.} Given the 3D scene, we need to follow the instruction to synthesize future motions (as shown in Fig.~\ref{fig:network} (a)). Besides the noised motion $x_t$, step $t$, and mask joints $\mathcal{J}$, we take the action $a$ encoded by a learnable codebook as our input.

Meanwhile, we need to follow the instruction to navigate to the target position or interact with the target object $o$. For locomotion, we will calculate the direction from current human pelvis to target position $\mathbf{d}=o-J_{1,0}\in\mathbb{R}^3$. The direction will be repeated for $K$ times where $K$ indicates the number of human joints. As for interaction, we will sample target human pose with skeleton joints $\bar{J}$, then we will calculate the direction for all joints $\mathbf{d}=\bar{J} - J_1\in\mathbb{R}^{K\times 3}$.
In addition, we clip the norm of $\mathbf{d}$ to a normalized range
\begin{equation}
    \mathbf{d}_n = \frac{\min(||\mathbf{d}||,1)+\epsilon}{||\mathbf{d||+\epsilon}}\mathbf{d},
\end{equation}
which provide directional hint $f_{dir}$ for motion control.

\subsubsection{Goal-Directed Human Scene Correlation.} For motion control, we mainly consider sub-task goal given by textual instructions and scene constrains. Here, we first model the correlation between human motion and goal human status (as shown in Fig.~\ref{fig:network} (b)). Specifically, we map directional hint feature $f_{dir}$ into $L_d$ tokens $f_{dir}^t$, which further provide the directional hint keys $K_{d,i}=W_{d,i}^kf_{dir}^t$ and values $V_{d,i}=W_{d,i}^vf_{dir}^t$. We further design to convert $f_{mot}$ into frame-wise queries $Q_{d,i}=W_{d,i}^qf_{mot}$. The goal-directed human motion feature can be derived by
\begin{equation}
    f_{mot}^{g}=\mathop{\text{concat}}\limits_{i}(\text{softmax}(\frac{Q_{d,i}K_{d,i}^T}{\sqrt{d_d}})V_{d,i})W_{o,d},
\end{equation}
where $W_{dir}^k$, $W_{dir}^v$, $W_{dir}^q$, and $W_{o,d}$ are learnable parameters. Furthermore, we map scene feature $f_{scene}$ into scene-related keys $K_{s,j}$ and values $V_{s,j}$, and model goal-directed human scene correlation via
\begin{equation}
    f_{mot}^{gs} = \mathop{\text{concat}}\limits_{j}(\text{softmax}(\frac{Q_{s,j}K_{s,j}^T}{\sqrt{d_s}})V_{s,j})W_{o,s},
\end{equation}
where $Q_{s,j}$ are frame-wise human motion queries given by $f_{mot}^g$. The goal-directed human scene correlation $f_{mot}^{gs}$ will be fed into the control branch.

\subsubsection{Human Motion Control.} The motion control branch is built proportionally to the main branch. Meanwhile, we employ multiple zero-initialized linear layers to inject control signal from the control branch into the main branch. As aforementioned, $f_{mot}^{gs}$ is fed into the control branch during training to ensure that the synthesized motion comply with the instruction requirements and scene constraints.

\subsection{Training and Inference}
Due to the scarcity of motion-in-scene datasets relative to motion capture datasets, SSOMotion adopts a decoupled training strategy in which the base diffusion model and control branch are trained separately on different datasets. 
\subsubsection{Fundamental Model Training.}
We follow the MDM \cite{tevet2023human} to train our base diffusion model. At each denoising step, the model predicts the clean motion sequence, and the training objective is to minimize the reconstruction loss between the predicted and ground-truth sequences.
To enable conditional generation, the model is trained with action labels and masked human joints as conditioning inputs. These forms of guidance are commonly available in standard motion capture datasets that do not include explicit scene context, making them well-suited for training the scene-agnostic base model.
\subsubsection{Control Branch Training.}
We train a separate control branch on a motion-in-scene dataset to incorporate scene awareness. Each scene is represented by the SSO to capture the semantic layout of the environment. 
During training, the parameters of the base diffusion and control branch are jointly optimized. The control branch receives the SSO representation along with the human movement direction as input and produces control signals that modulate the denoising process of the diffusion model. 
\subsubsection{Motion Synthesis.}
Our model supports multi-task motion synthesis in 3D scenes, including locomotion and different types of human-scene interactions. Given the textual instruction, we will decompose it in action-object pair $(a, o)$. The target object or position will be fed into the control branch along with the decomposed SSO to derive the control signal.
Meanwhile, the main branch will iteratively denoised the motion under the guidance of initial human status, action label, and control signal. In addition, we take advantage of the DIP~\cite{gong2026dip} to introduce scene constraints during denoising.
For long-term motion synthesis, our model can generate the future motion (consisting of $S$ frames) according to the historical motion constrains, with $H$ overlapping frames to ensure continuity in motion transition. The historical motion is represented in the form of masked subsequent human skeleton joints, which serve as the input to the main branch. The historical motion is updated by blending it with the synthesized motion. The motion synthesis process continues until all tasks are completed. Thus, our approach allows the model to produce coherent and goal-directed motion with indefinite length.

\section{Experiments}
\subsection{Datasets}\label{sec:dataset}

\subsubsection{Motion Datasets.}
In this work, we use AMASS~\cite{mahmood2019amass} dataset for base diffusion model training . Babel~\cite{punnakkal2021babel} provided action labels and the initial/final frames for several subsets of the AMASS dataset and HumanML3D~\cite{guo2022generating} supply additional sentence annotations and initial/final frames for more motion data in AMASS.
All motions are downsampled to 30 FPS and segmented into 30-frame clips with a 5-frame stride. Each motion clip is transformed based on the human pose in the first frame, aligning the initial pose to the origin with the body orientation facing the positive y-axis. These processed motion clips, along with each corresponding action labels, are then used to train the base motion diffusion model.

\subsubsection{Scene-aware Motion Datasets.}
We use the HUMANISE dataset \cite{wang2022humanise} for motion controller training, which aligns motion data from AMASS with scenes from ScanNet \cite{dai2017scannet}. We voxelize the dense scene point cloud with a voxel size of $4$cm to generate scene occupancy and apply a nearest-neighbor approach to propagate semantic labels from sparse annotations. The final scene semantic occupancy is stored in an compact version where coordinate–label pairs are used for all non-empty voxels.

\subsubsection{Scene Datasets.}
We evaluate the proposed method in two widely adopted dataset, namely PROX \cite{hassan2019resolving} and Replica \cite{straub2019replica}. By benchmark on the two dataset, we assess our model's long-term motion synthesis across diverse tasks. The experiments share the same motion diffusion model and pipeline, ensuring the method's generalization capability.

\subsection{Scene Navigation}
We take the cluttered scenes constructed by DIMOS~\cite{zhao2023synthesizing} for locomotion evaluation, where furniture from ShapeNet~\cite{chang2015shapenet} is randomly placed.
\subsubsection{Metrics.}
Following the DIMOS \cite{zhao2023synthesizing} metrics for scene navigation evaluation, we adopt four criteria: average distance from the final body position to the target point, foot contact score indicating the degree of ground contact, locomotion penetration score representing the percentage of body vertices within the walkable area, and the task completion time measured in seconds.

\subsubsection{Results.}
We report the results of motion synthesis for locomotion in Tab.~\ref{tab:locomotion}. The results demonstrate the proposed method can achieve cutting-edge performance in destination distance (0.02m), non-penetration score (0.95), and execution time (3.60s). The proposed method has inferior performance in foot contact score as we focus more on contact status of foot vertices rather than the inner human joints. 

\begin{figure}[h!]
    \centering
    \includegraphics[width=\linewidth]{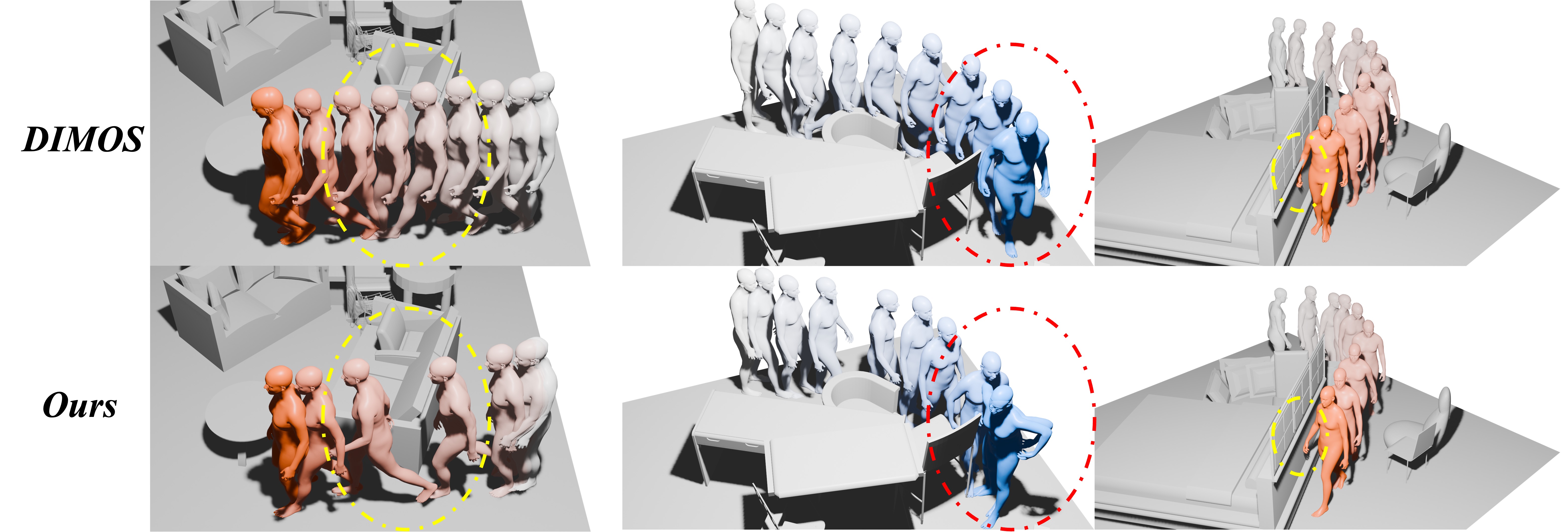}
    \caption{Visual comparison of locomotion synthesis between DIMOS and the proposed method.}
    \label{fig:locomotion}
\end{figure}

We present the visual results in Fig.~\ref{fig:locomotion} where the proposed method shows higher motion diversity and less collision with scenes.

\begin{table}
    \centering
    \begin{footnotesize}
    \begin{tabular}{lcccc}
        \toprule
         & avg. dist $\downarrow$ & ft. cont $\uparrow$ & loco. pene $\uparrow$ & time $\downarrow$\\
        \midrule
        SAMP & 0.14 & 0.84 & 0.94 & 5.97\\
        GAMMA & 0.03 & 0.94 & 0.94 & 3.87\\
        DIMOS & 0.04 & \textbf{0.99} & \textbf{0.95} & 6.43\\
        OmniControl & 0.04 & 0.76 & 0.93 & \textbf{2.43}\\
        Ours & \textbf{0.02} & 0.90 & \textbf{0.95} & 3.60\\
        \bottomrule
    \end{tabular}
    \end{footnotesize}
    \caption{Evaluation of motion synthesis on locomotion task. The up/down arrows ($\uparrow$/$\downarrow$) indicate higher/lower is better. Metrics with best performance are annotated in boldface.}
    \label{tab:locomotion}
\end{table}

\subsection{Human-Scene Interaction}
Following previous work~\cite{zhao2023synthesizing}, we take the scenes consisting of specific $10$ furniture from the ShapeNet~\cite{chang2015shapenet} for interaction evaluation.
\begin{figure}[h!]
    \centering
    \includegraphics[width=\linewidth]{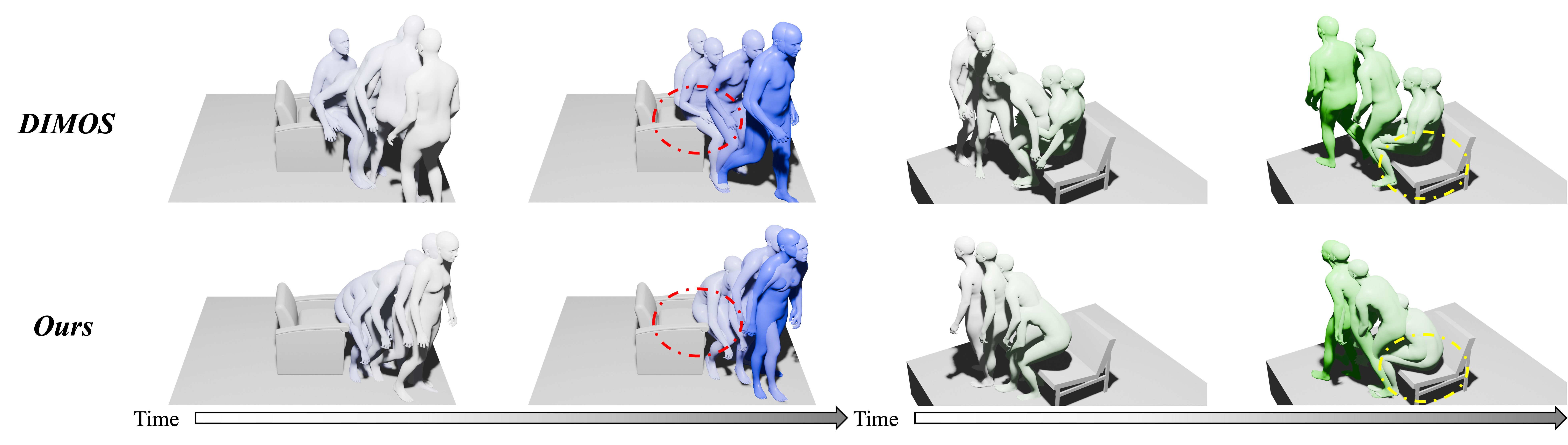}
    \caption{Visual results given by DIMOS and the proposed method for sitting action.}
    \label{fig:sitting}
\end{figure}

\subsubsection{Metrics.}
We take three metrics to evaluate the performance of human-scene interaction, including task completion time, mean human mesh vertex penetration based on vertex SDF values with scene, and maximum penetration observed over time.

\subsubsection{Results.}
We report the results of motion synthesis for scene interaction in Tab.~\ref{tab:interaction}. It can be seen, the synthesized motion given by the proposed method can execute the instruction more rapidly with minimum scene penetration. 

\begin{figure}[h!]
    \centering
    \includegraphics[width=\linewidth]{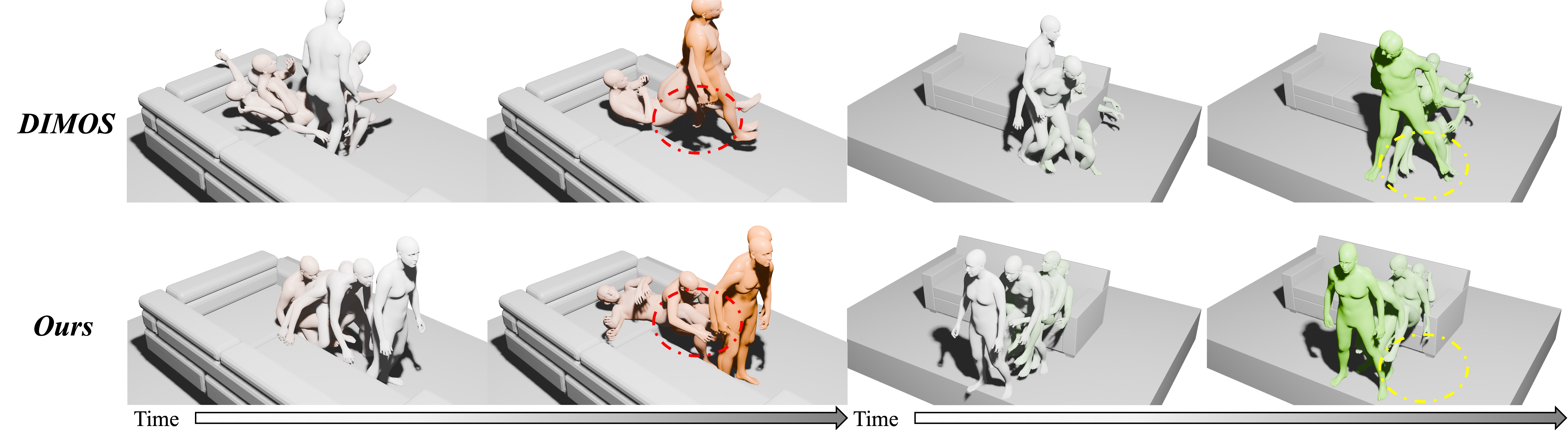}
    \caption{Visualization of lying motions synthesized by DIMOS and the proposed method. }
    \label{fig:lying}
\end{figure}

\begin{table}
    \centering
    \begin{footnotesize}
    \begin{tabular}{lcccc}
        \toprule
        &action & time $\downarrow$ & pene. mean $\downarrow$ & pene. max $\downarrow$ \\
        \midrule
        SAMP &sit& 8.63 & 11.91 & 45.22 \\
        DIMOS &sit& 4.09 & 1.91 & 10.61\\
        Ours &sit& \textbf{3.45} &  \textbf{1.83}& \textbf{6.78} \\
        \midrule
        SAMP &lie & 12.55 & 44.77 & 238.81 \\
        DIMOS &lie & 4.20 & 9.90 & 44.61 \\
        Ours &lie &  \textbf{3.69} &  \textbf{5.74} & \textbf{40.48} \\
        \bottomrule
    \end{tabular}
    \end{footnotesize}
    \vspace{-0.1cm}
    \caption{Evaluation of motion synthesis on interaction tasks. The up/down arrows ($\uparrow$/$\downarrow$) indicate higher/lower is better. The best results are shown in boldface.}
    \vspace{-0.1cm}
    \label{tab:interaction}
\end{table}

\subsection{Long-term Motion Synthesis}
We attempt to synthesize long-term human motion in 3D scenes following consequent textual instructions. We utilize the COINS~\cite{zhao2022compositional} to populate the human in target position following textual instruction, which further guide goal-directed motion synthesis.

\subsubsection{Metrics.}
We conducted a comprehensive user study to obtain a more accurate evaluation of long-term motion synthesis performance. Participants were asked to give scores to the generated motions without prior knowledge of the corresponding method. Each participant assessed four metrics on all generated motion samples, including naturalness, diversity, plausibility, and goal achievement.

\subsubsection{Results.} 
\begin{figure}[h!]
    \centering
    \includegraphics[width=\linewidth]{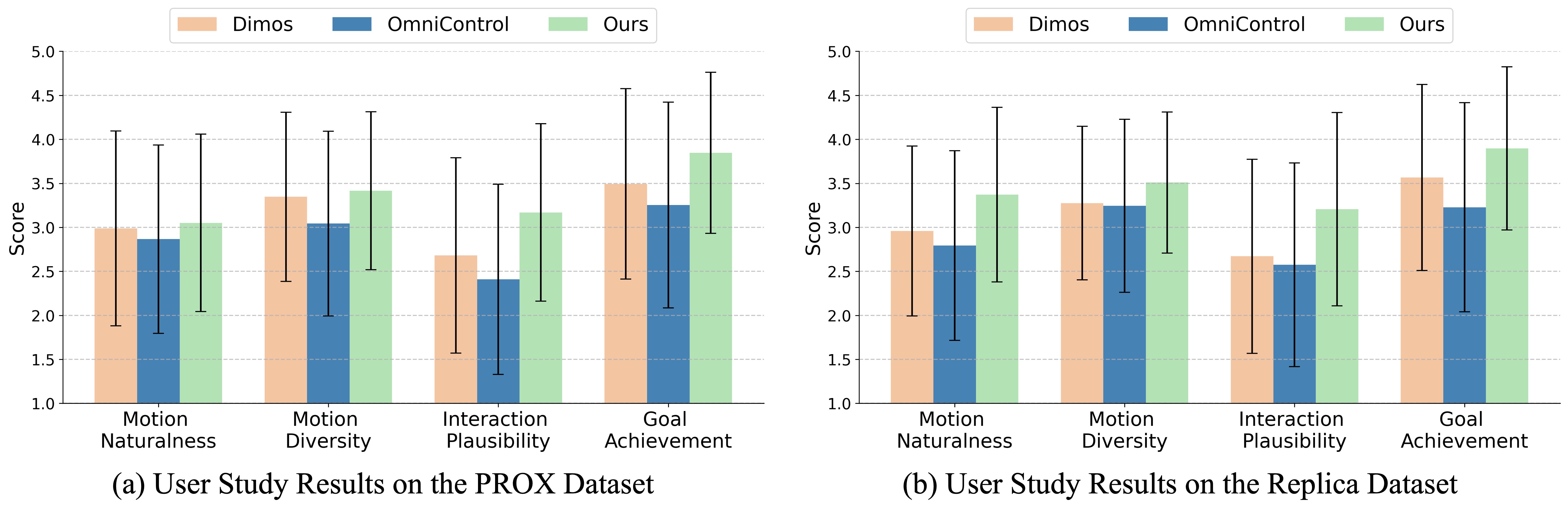}
    \caption{(a) and (b) present the scores of competitive methods across four aspects in the user study. Higher scores indicate better performance.}
    \label{fig:user_study}
\end{figure}

Finally, we collect $4,080$ ratings from $17$ participants and report the results in Fig.~\ref{fig:user_study}. It can be observed that our method achieves optimal performance across all aspects. 

\begin{figure*}[h!]
    \centering
    \includegraphics[width=\linewidth]{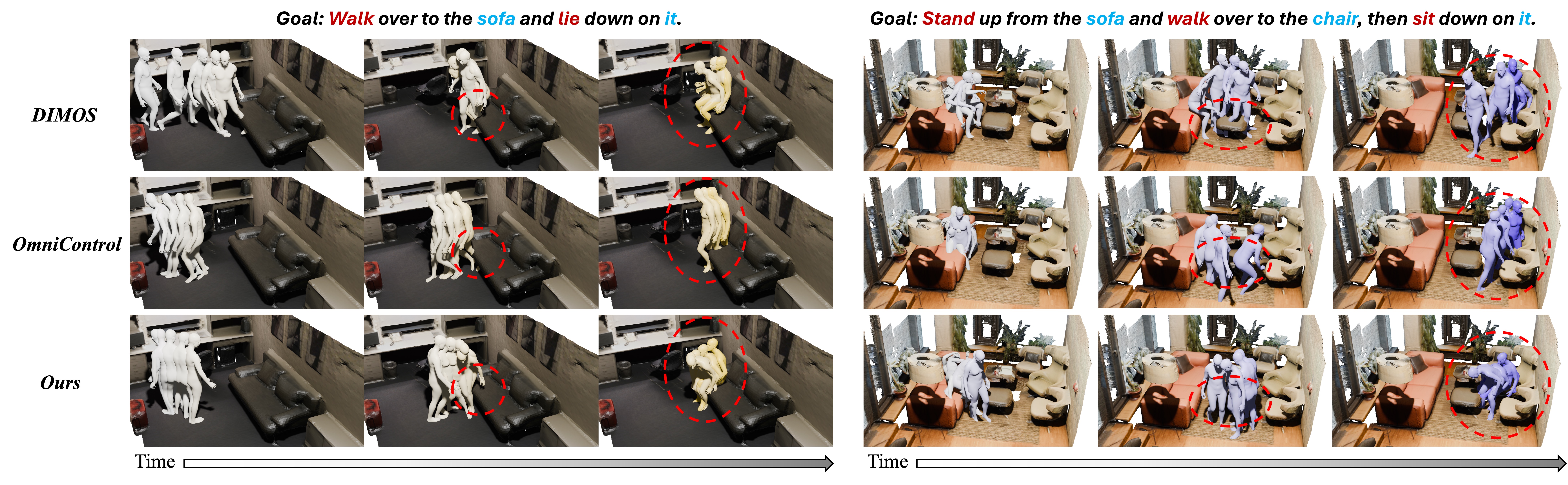}
    \caption{Visual comparison of different methods for instruction-based motion synthesis in 3D scenes from PROX dataset.}
    \label{fig:prox}
\end{figure*}

\begin{figure*}
    \centering
    \includegraphics[width=\linewidth]{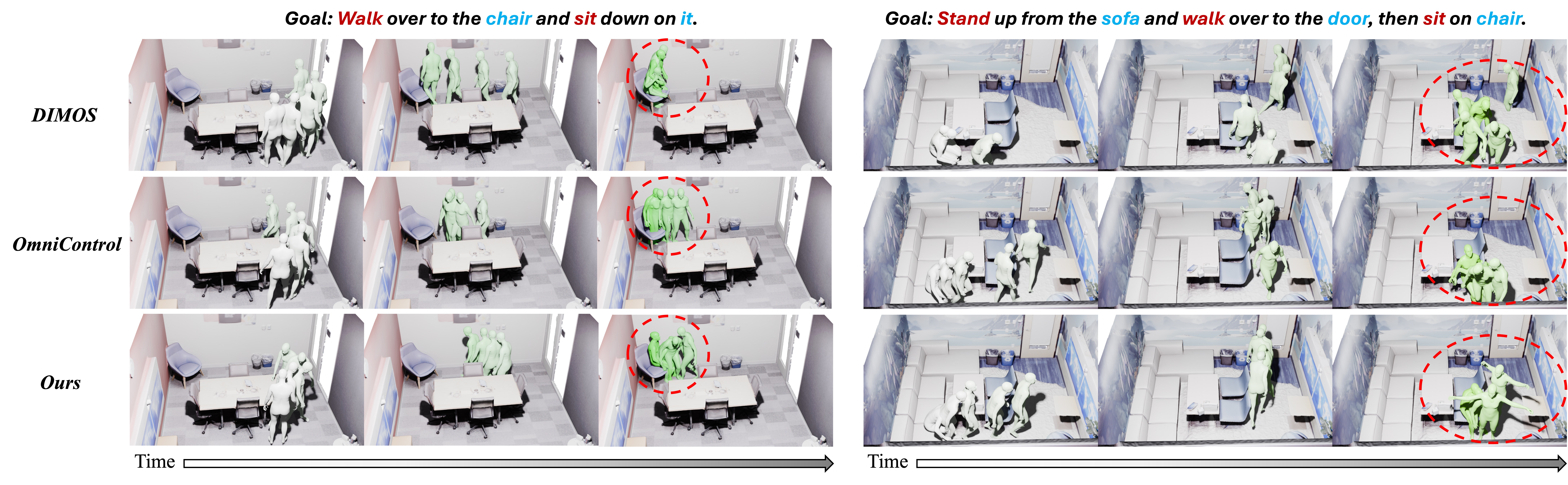}
    \caption{Visualization of synthesized human motion given by different method in 3D scenes from the Replica dataset.}
    \label{fig:replica}
\end{figure*}

We also present the visual differences between competitive methods in Fig.~\ref{fig:prox} and Fig.~\ref{fig:replica} for motion synthesis in scenes from the PROX and Replica dataset.

\subsection{Ablation Study}
In this part, we mainly analyze the computational cost of scene comprehension and human scene correlation in scene-aware human motion synthesis.

\begin{table}[]
    \centering
    \begin{tabular}{ccc}
         \toprule
         BT Decomp.&  Dim. Red.& Comp. Cost (GFLOPs)\\
         \midrule
         & & 20783.78\\
         \checkmark &  & 421.73\\
          & \checkmark & 21.21\\
         \checkmark & \checkmark & 0.49\\
         \bottomrule
    \end{tabular}
    \caption{Computational Cost of the Scene Comprehension and Motion Scene Correlation for only $1$ sample. 
    }
    \label{tab:cost}
\end{table}

\subsubsection{Computational Cost.}
In the proposed method, we decompose the scene semantic occupancy into bi-directional tri-plane RGB-DS map for scene comprehension and motion control. Before semantic feature map construction, we introduce a shared linear layer for dimensionality reduction of semantic feature from CLIP textual encoder.

We report the computational cost (batch size equals $1$) of (1) Neither Bi-directional Tri-plane Decomposition (BT Decomp.) nor Dimensionality Reduction (Dim. Red.), (2) only ``BT Decomp.'', (3) only ``Dim. Red.'', and (4) both ``BT Decomp.'' and ``Dim. Red'' in Tab.~\ref{tab:cost}. It can be seen, such design can significantly reduce computational cost while preserve the scene semantic occupancy information. Without the SSO Bi-directional Tri-plane Decomposition and semantic feature Dimensionality Reduction, employing high-resolution SSO for scene comprehension in motion synthesis is computationally prohibitive.

\section{Conclusion}
In this paper, we present SSOMotion, an effective framework for human motion synthesis in 3D scenes. 
We adopt a novel approach to integrate semantic and structural scene understanding by leveraging scene semantic occupancy. Through a bi-directional tri-plane decomposition of the SSO and applying a dimensionality reduction on CLIP-based semantic encoding, our model is capable to efficiently capture fine-grained scene semantics while minimizing the computational overhead.
We then validate the effectiveness and generalization ability of the proposed SSOMotion through extensive experiments on synthesized scenes from DIMOS with ShapeNet furniture, and cluttered scenes from PROX and Replica datasets. 

Despite its effectiveness, SSOMotion currently has certain limitations and future work should focus on optimizing the efficiency and extending the framework to support real-time deployment in dynamic environments.

\section*{Acknowledgements}
This work is supported by the National Natural Science Foundation of China (Grant No. 62222602, 62302167, U23A20343, 62502159), Natural Science Foundation of Shanghai (Grant No. 25ZR1402135), Shanghai Sailing Program (Grant No. 23YF1410500), Young Elite Scientists Sponsorship Program by CAST (Grant No. YESS20240780), the Chenguang Program of Shanghai Education Development Foundation and Shanghai Municipal Education Commission (Grant No. 23CGA34), Natural Science Foundation of Chongqing (Grant No. CSTB2023NSCQ-JQX0007,CSTB2023NSCQ-MSX0137, CSTB2025NSCQ-GPX0445), Open Project Program of the State Key Laboratory of CAD\&CG (Grant No. A2501), Zhejiang University.

\bibliography{aaai2026}

\appendix

\section{Implementation Details}
\subsection{Data Pre-processing}
We take an unified format for motion data across different datasets~\cite{mahmood2019amass,punnakkal2021babel,guo2022generating,wang2022humanise}. Specifically, we segment motions into clips of 30 frames at 30 FPS. We employed a sliding window approach with a stride of 5 when extracting motion clips longer than 30 frames.

Each motion clip is spatially aligned according to the pose of the first frame. Namely, the root joint (pelvis) is translated to the origin $(0,0,0)$, and the global orientation is adjusted horizontally such that the body face the positive y-axis within the yz-plane. These normalized motion clips, paired with their corresponding action labels, were used to train the motion diffusion model. Additionally, randomly-masked skeleton joints are extracted utilizing SMPL-X human model~\cite{pavlakos2019expressive} to serve as positional guidance during training. 

\subsection{Bi-directional Tri-plane SSO}
All scenes from HUMANISE~\cite{wang2022humanise}, PROX~\cite{hassan2019resolving}, and Replica~\cite{straub2019replica} are converted into Scene Semantic Occupancy $\mathcal{S}\in\mathbb{R}^{N\times 8}$ utilizing 3D scene models and semantics. We define the voxel size to be $4$cm. The color of each voxel is determined by the average value of the points within the voxel, while the corresponding semantics are determined by the mode of the semantics of the points within the voxel. Finally, each element $\mathcal{S}_i$ in $\mathcal{S}$ indicate an occupied voxel, consisting of $xyz$ coordinate, $rgba$ color and semantic label $s$.

For motion model training and motion synthesis within scenes, we need to derive the local SSO according to the scene SSO and human pose at the first frame. We construct a new local coordinate where the origin $(0,0,0)$ is located at the pelvis of initial human, positive $x$-axis is directed to the subject's right, and $z$-axis is aligned with the world coordinate system. 
In the local coordinate, we employ $101\times101\times81$ sensors within a $4\text{m}\times4\text{m}\times3.2\text{m}$ bounding box. Then, we change the coordinates of sensors to world coordinate and derive the local SSO $\mathcal{S}_l \in \mathbb{R}^{101 \times 101 \times 81 \times 5}$ by grid sampling.

Finally, we decompose the local SSO into bi-directional tri-plane format where ceilings are ignored. Specifically, RGB values and semantic labels are projected along the $\pm x$-axis, $\pm y$-axis, and $z$-axis. In addition, the scene depth map is derived from the scene occupancy, forming the semntic occupancy map $\mathcal{O}_{i,j}$ mentioned in the main manuscript.

\subsection{Motion Scene Correlation}
The scene hint $\mathcal{I}_{scene}$ aforementioned, along with frame-wise motion feature $f_{mot}\in\mathbb{R}^{30\times 512}$, directional hint $\mathcal{I}_{di}\in\mathbb{R}^{66}$ are taken as input for the Motion Scene Correlation module. The directional hint will be encoded through MLPs to obtain the directional hint feature $f_{dir}\in\mathbb{R}^{8\times 64}$. Meanwhile, the scene hint will also be mapped into scene feature $f_{scene}\in\mathbb{R}^{8\times64}$ via parallel MLPs for RGB-D and Semantics encoding. The latent dimension for cross attention is set to $512$ for Eq. 8 and Eq. 9 of the main manuscript.

\subsection{Loss Functions}
We utilize a series of motion reconstruction loss functions for the diffusion model training. The overall loss function is as follows:

\begin{equation}
\mathcal{L}_{\text{total}} = \mathcal{L}_{\text{rot}} +
\mathcal{L}_{\text{trans}} +
\mathcal{L}_{\text{vel}}
\end{equation}

\subsubsection{Reconstruction Loss}
The diffusion model is designed to reconstruct the whole motion. So, our primary loss is designed to supervise the joint rotation and human translation reconstruction:

\begin{equation}
\mathcal{L}_{\text{rot}} = M\cdot ||\hat{x}_{0}^{rot}-x_{0}^{rot}||^2,
\end{equation}

\begin{equation}
\mathcal{L}_{\text{trans}} = M\cdot ||\hat{x}_{0}^{trans}-x_{0}^{trans}||^2,
\end{equation}
where $\hat{x}_0^{rot},x_{0}^{rot}$ indicates the reconstructed and GT joint rotation and $\hat{x}_0^{trans},x_0^{trans}$ means the reconstructed and GT human translation. $M$ is a frame-wise mask for the motion sequence.

\subsubsection{Velocity Loss}
In addition, we also attempt to constrain the velocity of skeleton joint to be close to the Ground Truth. We derive the target human joint ($\mathcal{J}$) through SMPL-X human model~\cite{pavlakos2019expressive}. We then calculate the velocity of each joint through
\begin{equation}
         v = ||\mathcal{J}[1:S] - \mathcal{J}[0:S-1])||_2\times \nu 
\end{equation}
where $\nu$ indicate the FPS. The joint velocity for reconstructed motion $\hat{v}$ can be derived in similar way. By now, we can supervised the joint velocity in motion reconstruction.

\begin{equation}
\mathcal{L}_{\text{vel}} = 
M_{1:}\cdot ||\hat{v}-v||^2,
\end{equation}
where $M_{1:}$ indicated the frame-wise mask from the second frame.

\subsection{Computing Infrastructure}
All the experiments can be conducted on a single RTX A6000 with 64G RAM. The versions of
relevant software libraries can been seen in the supplementary code.

\section{More Visual Results}

\begin{figure*}
    \centering
    \includegraphics[width=\linewidth]{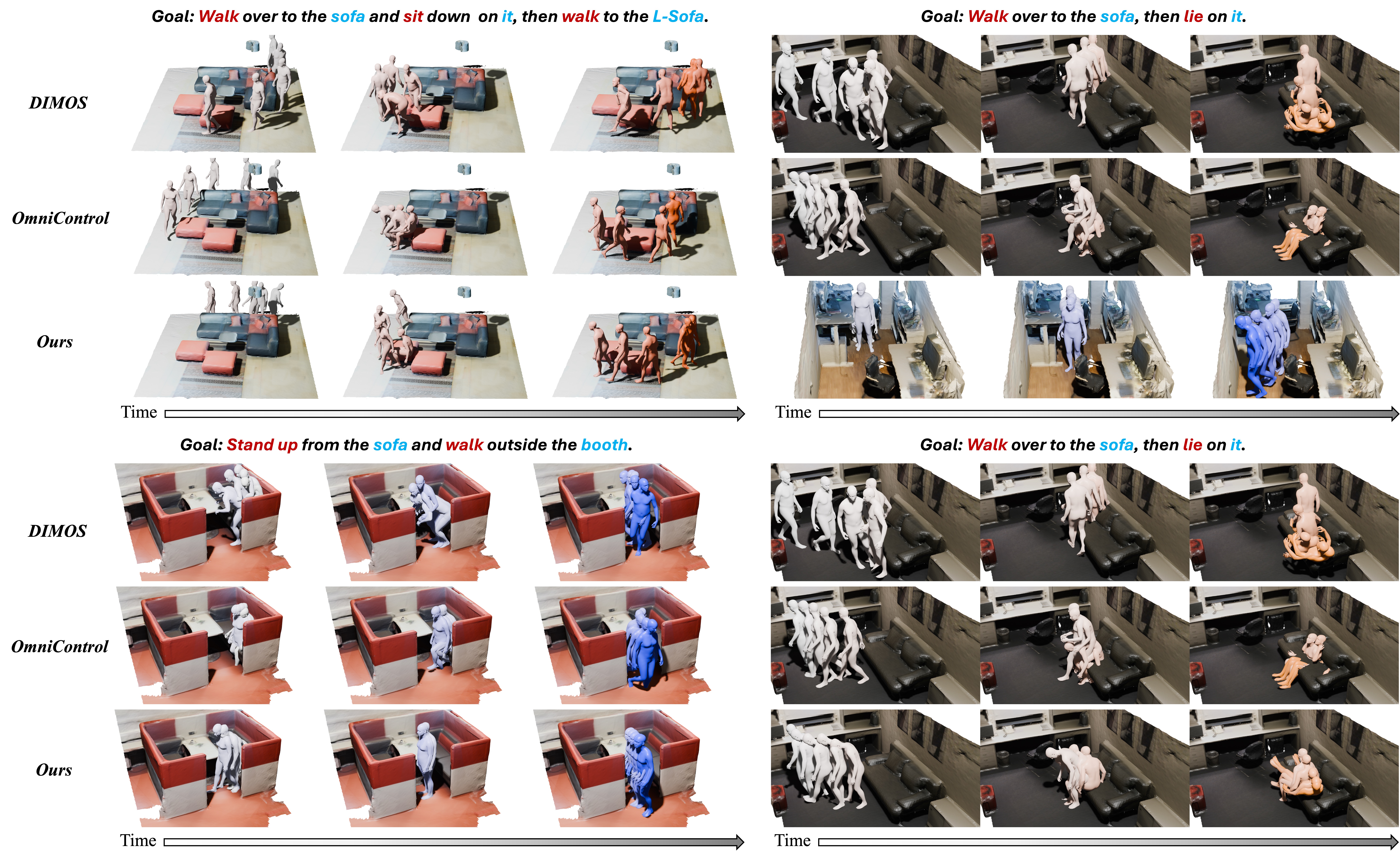}
    \vspace{-0.5cm}
    \caption{More visual results of motion synthesis in 3D scenes from PROX dataset.}
    \label{fig:prox_supp}
\end{figure*}

\begin{figure*}
    \centering
    \includegraphics[width=\linewidth]{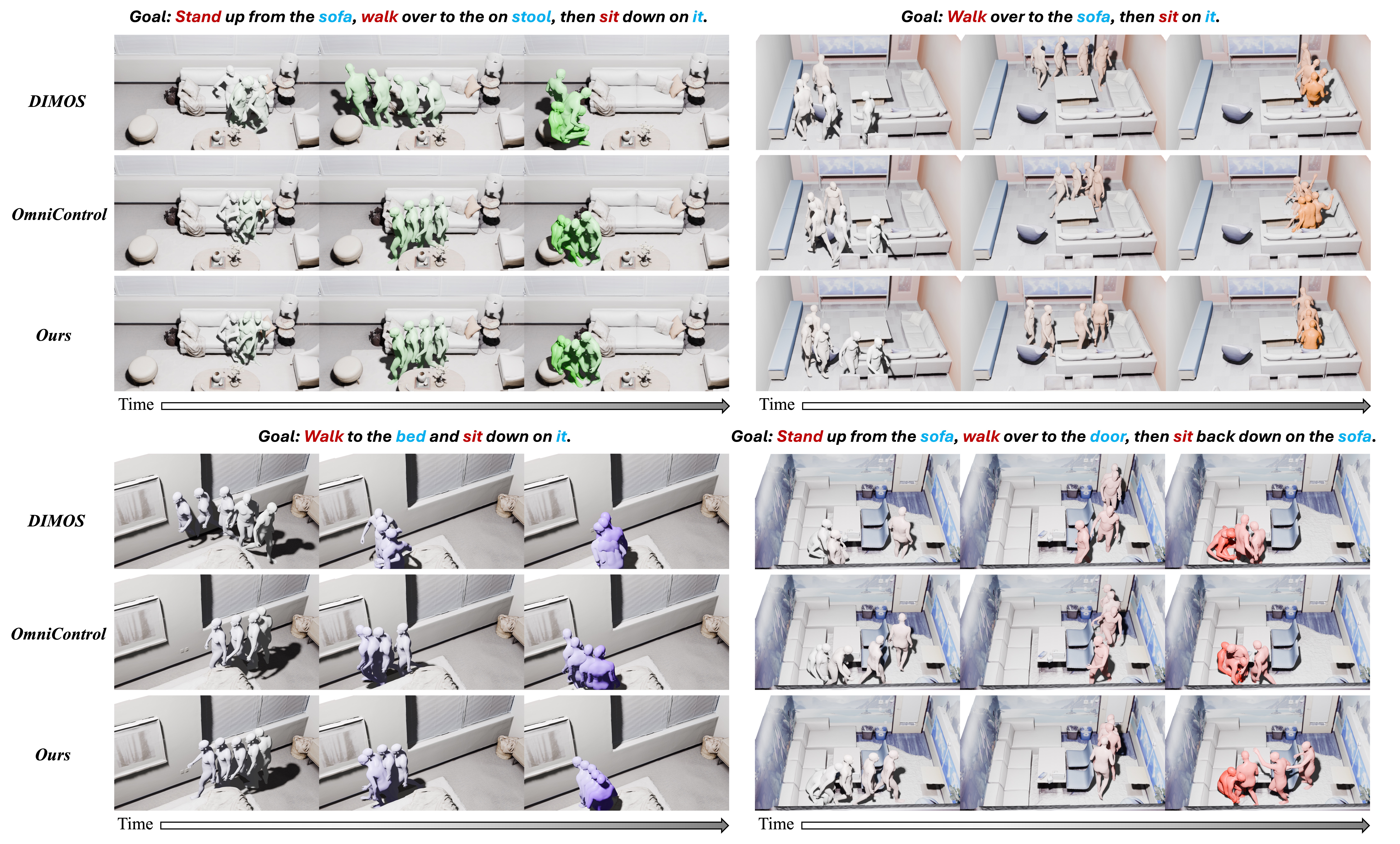}
    \caption{More visual results of motion synthesis in 3D scenes from Replica dataset.}
    \vspace{-0.5cm}
    \label{fig:replica_supp}
\end{figure*}

In this section, we provide more visual results of competitive methods on long-term motion synthesis in scenes from PROX and Replica. Fig.~\ref{fig:prox_supp} presents more visual results in 3D scenes from PROX dataset, and Fig.~\ref{fig:replica_supp} shows additional visual results in scenes from Replica dataset.

\section{More Ablation Studies}

\subsection{Spatial Semantic Information} 

\begin{figure}[h]
    \centering
    \includegraphics[width=\linewidth]{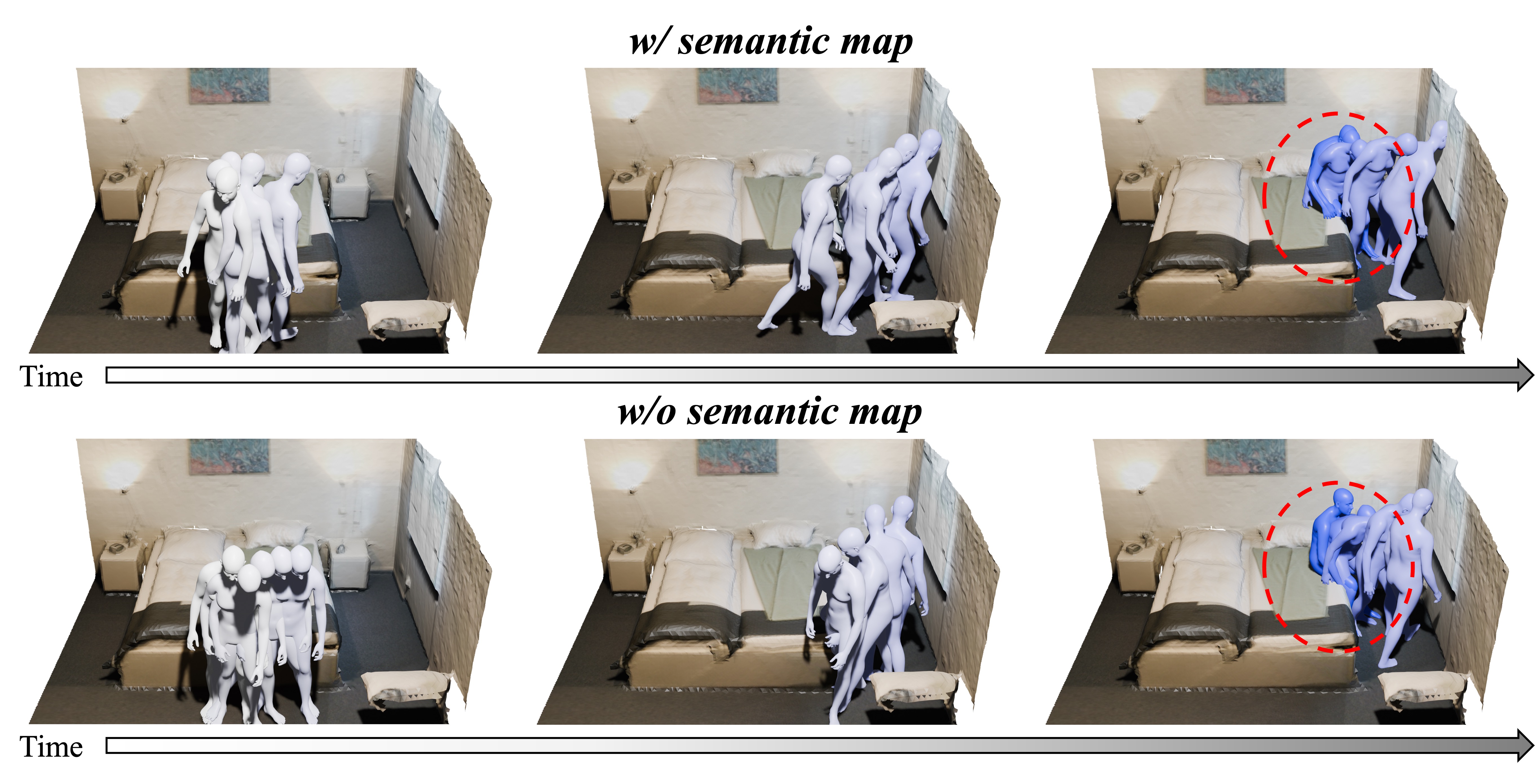}
    \caption{Visual comparison of the proposed method with/without spatial semantic information.}
    \label{fig:semantic}
\end{figure}

Different from typical grid occupancy sensor, we take a unified scene semantic occupancy for scene representation. To analyze the contribution of semantic information, we attempt to train the motion controller and evaluate the performance of motion prediction with or without tri-plane semantic map. The visual comparison is presented in Fig.~\ref{fig:semantic}. As shown in this figure, motions synthesized with/without the semantic map can both satisfy the command instruction. However, motion generated under the guidance of spatial semantics can better recognize the area for sitting. 

\subsection{Motion Prediction}
\begin{figure*}[ht]
    \centering
    \includegraphics[width=\linewidth]{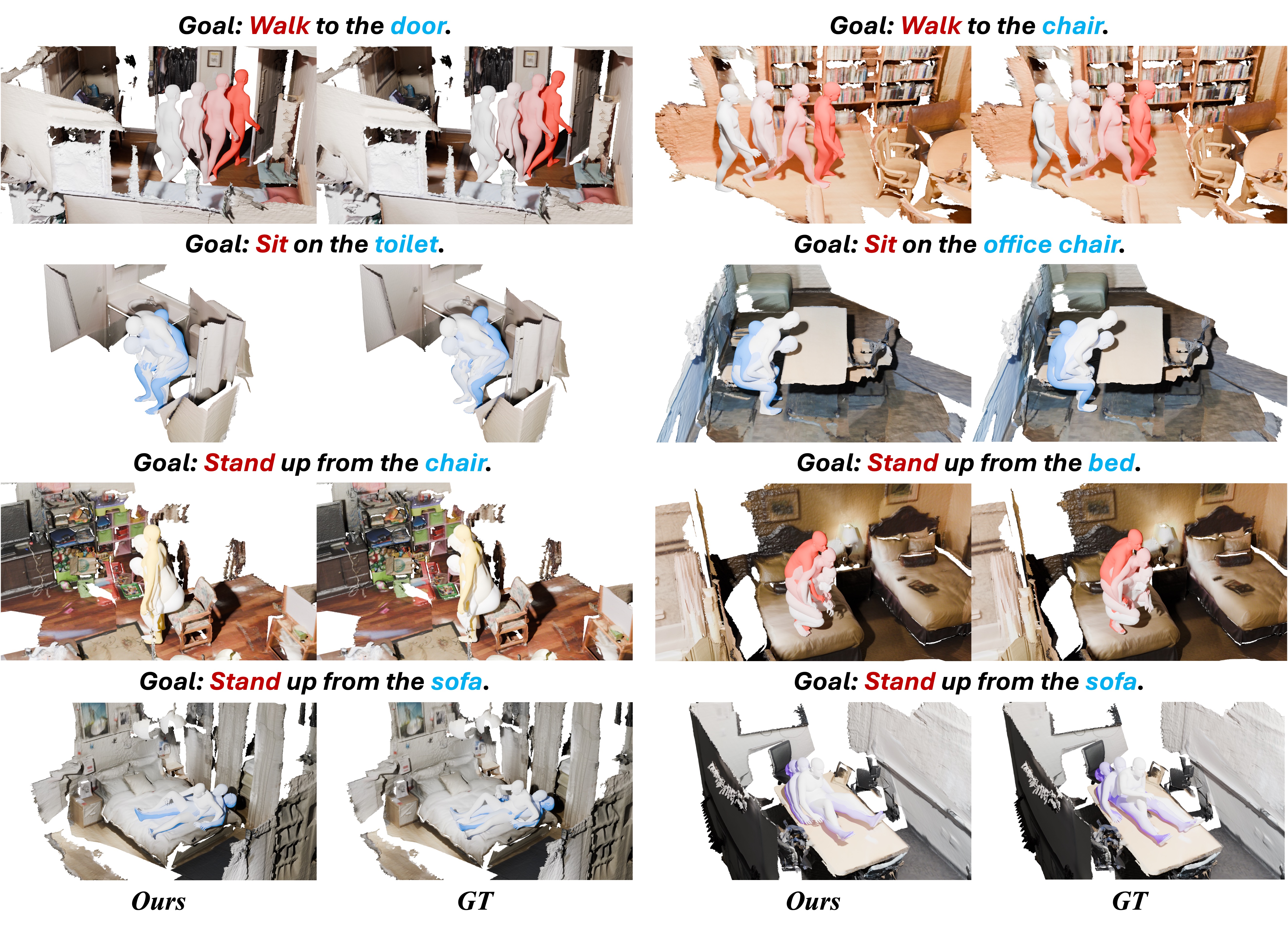}
    \caption{Visual results of atomic motion prediction on the test set of the HUMANISE dataset.}
    \label{fig:humanise_supp}
\end{figure*}

In addition, we apply the proposed scene-aware motion synthesis in motion prediction. The evaluation is conducted on the test set of HUMANISE~\cite{wang2022humanise}. Given the 3D scenes and initial human poses, we need to predict the motion in the future $30$ frames. We report the results in Tab.~\ref{tab:prediction}. 
In Fig.~\ref{fig:humanise_supp}, we also visualize the results of atomic motion prediction within $30$ frames on the test set of the HUMANISE. It can be seen, our SSOMotion can accurately anticipate the future motion according to the goal direction and scene semantic structure. 

\begin{table}[]
    \centering
    \begin{tabular}{ccc}
         \toprule
         MPJPE (m) $\downarrow$ & MPVPE (m) $\downarrow$ &  Goal Dist. (m) $\downarrow$\\
         \midrule
         0.095 &  0.118 & 0.045\\
         \bottomrule
    \end{tabular}
    \caption{Results of the proposed method on the motion prediction task in the HUMANISE dataset. Mean Per Joint Position Error (MPJPE), Mean Per Vertex Position Error (MPVPE), and Goal Distance measured in meters are taken as the evaluation metrics. $\downarrow$ indicates lower is better. }
    \label{tab:prediction}
\end{table}

\section{Discussion}
As shown in the experimental results and user study, the proposed method shows good performance and high generalization ability due to the unified SSO representation.

On the other side, there is still limitations in the proposed method. The scene is perceived through Scene Semantic Occupancy in grid, resulting in deviation from original scene mesh, thus slight collision and hovering ($4$cm) may occasionally occur. The proposed method is specifically designed for single human motion synthesis. Crowd generation may benefit from the proposed method, but the length of motion synthesized within single-round may be abbreviated for higher responsiveness.

\end{document}